\documentstyle[elsart12,named,lingmacros]{article} 
\title{Discourse Coherence and Shifting Centers in Japanese Texts}
\author{
Masayo Iida \\
Fujitsu Software Corporation\\
masayo@ossi.com
}

\date{}

\def\longex#1#2#3#4#5#6#7{\begin{tabular}[t]{@{}l}
 \begin{tabular}[t]{@{}*{#1}{l@{\ \ }}}
 #3\\ #5 \end{tabular}\vspace{.5ex}\\
 \begin{tabular}[t]{@{}*{#2}{l@{\ \ }}}
 #4\\ #6\\[.5ex] \multicolumn{#2}{@{}l@{}}{\begin{tabular}[t]{@{}l@{}}
 #7 \end{tabular}} \end{tabular} \end{tabular}}

\begin{document}           
\bibliographystyle{named}  
\maketitle                 


\begin{abstract}

In languages such as Japanese, the use of {\it zeros}, unexpressed
arguments of the verb, in utterances that shift the topic involves a
risk that the meaning intended by the speaker may not be transparent
to the hearer. However, this potentially undesirable conversational
strategy often occurs in the course of naturally-occurring discourse.
In this chapter, I report on an empirical study of 250 utterances with
{\it zeros} in 20 Japanese newspaper articles. Each utterance is
analyzed in terms of centering transitions and the form in which
centers are realized by referring expressions. I also examine lexical
subcategorization information, and tense and aspect in order to test
the hypothesis that the speaker expects the hearer to use this
information in determining global discourse structure. I explain the
occurrence of {\it zeros\/} in {\sc retain} and {\sc rough-shift}
centering transitions, by claiming that a {\it zero\/} can only be
used in these cases when the shift of centers is supported by
contextual information such as lexical semantics, tense and aspect,
and agreement features.  I then propose an algorithm by which
centering can incorporate these observations to integrate centering
with global discourse structure, and thus enhance its ability for
non-local pronoun resolution.

\end{abstract}

\section{Introduction}

Centering Theory is a computational model of discourse interpretation
that examines the relationship between attentional state, the form of
referring expressions, and the control of inferential processes. These
goals have led to its application to the study of unexpressed
arguments (henceforth {\it zeros\/}) in topic-oriented languages like
Japanese, in which salient entities, recoverable by inference in a
given context, are freely omitted. Centering predicts the preferred
interpretation of {\it zero\/}s in situations in which the antecedent
of a {\it zero\/} was realized as a center in the previous discourse.

Previous work argues that both syntactic and discourse factors
associated with potential antecedents determine the preferred
interpretation of {\it zero\/}s
\cite{Kuno73,Kameyama85,WIC90,Iida92,WIC94}. For example, a discourse
entity realized as a subject is more likely to serve as the antecedent
of a {\it zero\/} than a discourse entity realized as a object.
Walker et al. incorporated certain discourse features into centering
with their proposed rule of {\sc zero topic assignment} (henceforth
{\sc zta}). This proposal was motivated by the observation that a {\it
zero\/} that was previously the center of attention (i.e., Cb) is
easily understood as the continuing center even if it is expressed in
a syntactically less salient argument position.\footnote{The salience
of the subject has also been observed in various syntactic phenomena
such as extraction and binding. See Walker et al.~(1994) for the
discussion of salience among the arguments of the verb in Japanese.}
For example, a zero object in a given utterance such as \ex{1}c below,
is the topic because it was the Cb in the previous utterance. As the
topic, the discourse entity realized in object position is ranked
higher on the Cf list than the discourse entity realized as the
subject. This explains the preferred interpretation of the subsequent
utterance, (\ex{1})d in this case, as will be discussed in more detail
below.

\eenumsentence{\item[a.]
\shortex{10}{Hanako &wa &siken &o &oete,&kyoositu&ni&modorimasita.}
          {Hanako&{\sc top/subj} &exam &{\sc obj} &finish &classroom &to
&returned}
          {{\it Hanako returned to the classroom, having finished her exam.}}

\item[b.]
\shortex{6}{0  &hon &o &locker &ni &simaimasita.}
           {{\sc subj} &book &{\sc obj} &locker &in &took-away}
           {{\it She put her books in the locker.}}

\item[c.]
\shortex{10}{Itumo &no &yooni &Mitiko &ga &0&deki&o&tazunemasita.}
{always& &like &Mitiko&{\sc subj} &{\sc obj2} &result &{\sc obj}
&asked}
{{\it Mitiko, as usual,  asked (Hanako) how she did.}}

\item[d.]
\shortex{10}
{      0 &   0 &zibun&no&tokenakkatta&mondai&o &misemasita.}
{{\sc subj} & {\sc obj2} &self&{\sc gen}&solve-could-not&problem&{\sc
obj}&showed}
{{\it (Hanako) showed (Mitiko) the problems which she could not
solve. }}
\label{intro-examp}
}

In order to further test the feasibility of {\sc zta} and to examine
strategies for keeping track of centers, this chapter examines the
distribution of {\sc zeros} in naturally occurring Japanese newspaper
texts. Two initial hypotheses  about the use of  {\it
zeros\/} are given in \ex{1} and \ex{2}:

\eenumsentence{\item[\ ]
{\bf Hypothesis-1} \\
{\it Zeros\/} are used to {\sc continue} the center.}

\eenumsentence{\item[\ ]
{\bf Hypothesis-2} \\
Full {\sc
np}s are used to {\sc shift}  the center.}

These hypothesis are similar to one tested for Italian in (Di Eugenio,
this volume).\footnote{Di Eugenio's hypothesis says, ``Typically a
null subject signals a {\sc continue}, and a strong pronoun a {\sc
retain} or a {\sc shift}.}  I report on an empirical study based on
250 utterances from a corpus of 20 Japanese newspaper articles.  Each
utterance is analyzed in terms of centering transitions and the form
in which centers are realized by referring expressions. I also examine
lexical subcategorization information, and tense and aspect in order
to test the hypothesis that the speaker expects the hearer to use this
information in determining global discourse structure.  Figure
\ref{trans-state-fig} summarizes the findings on the distribution of
centering transitions with respect to form of referring expression
used in the utterance.

\begin{figure}[htb]
\begin{tabular}{|r|c|c|c|c|}
\hline
& & & &  \\
& {\sc continue} & {\sc \ \ \ retain\ \ \ } & {\sc smooth-shift} & {\sc
rough-shift}   \\
\hline
& & & & \\
{\it with\/} {\sc zero\/} & 76 & 3 & 34 & 23   \\ \hline
& & & & \\
{\it without\/} {\sc zero\/} & 7 & 39 & 9 & 35   \\ \hline
& & & & \\
total & 83 & 42 & 42 & 58  \\ \hline
\end{tabular}
\caption{Distribution of Centering  Transitions and Zeros
in Japanese newspaper texts}
\label{trans-state-fig}
\end{figure}


The hypothesis in (\ex{-1}) is confirmed by the distribution of {\sc
continue} transitions in Figure \ref{trans-state-fig}, as compared to
the other transitions combined $\chi^2$ = 53.932, {\it p\/}\ $<$\
.001). In {\sc continue} transitions, {\it zeros\/} are strongly
preferred to {\sc np}s: among {\sc continue} transitions, 76 cases
appears with {\it zero\/} and only 7 cases without {\it zero\/}, while
other transitions preferentially realize centers by {\sc np}s: there
are 60 cases with {\it zeros\/} and 83 cases without {\it zeros\/}.

Note that  Hypothesis-1 predicts as a corollary that discourse
entities ranked higher in the Cf ranking would tend to be
realized by {\it zeros\/}.  The preference of {\it zeros\/} in {\sc
continue} transitions proves this tendency and  provides additional
 support for Walker etal's rule of {\sc  zero topic assignment}.

However, the second hypothesis in \ex{0} is disconfirmed: while the frequency
of full NPs is greater (83) than zeros (60), full {\sc
np}s are {\bf not} always used to shift the center, and zeros frequently are.
 The distribution
of centering transitions in figure \ref{trans-state-fig} shows that a
shift of attentional state is abundant in naturally occurring
discourse, as seen by the frequency of {\sc retain} and {\sc rough-shift},
which the centering algorithm prefers the least \cite{BFP87}.  In the Japanese
data
examined here, these transition states are identified when a {\sc
zero} cannot take the current center of attention, the Cb, as its
antecedent.

Thus, what needs to be explained is the occurrence of {\it zeros\/} in
these  transitions in which the Cb changes, where it may be
difficult for the hearer to determine which discourse entity is
realized by the {\it zero}.  How is discourse coherence preserved when
two adjacent utterances are not locally coherent? In the transition
state of {\sc retain}, the rule of {\sc zta} makes it is possible to
avoid shifting the Cb, but in a {\sc rough-shift} transition, there is
no link to the prior utterance, and the Cb must shift.\footnote{What I
call a {\sc rough shift} in this chapter is elsewhere called a {\sc no
cb} transition. That is, there is no Cb as no entity from the
Cf(U$_{n-1}$) is realized in the current utterance.}

The main focus is of this chapter is to study the relation of local
and global structure in discourse, by exploring the strategies that a
speaker uses to reduce the hearer's inference load and make the flow
of discourse coherent when the antecedent of a {\it zero\/} is not
realized  in the immediately preceding utterance. In section
\ref{zta-sec}, I discuss in more detail how centering works in
Japanese and the rule of {\sc zero topic assignment}\cite{WIC94}. Then
in section \ref{shift-sec}, I show how cues such as lexical semantics
and tense and aspect can be used to interpret zeros in utterances that
realize {\sc rough shift} transitions. On the basis of this analysis,
section \ref{integ-sec} sketches an algorithm for integrating
centering with global focus, and finally section \ref{discuss-sec}
summarizes the contributions of the chapter.

\section{Zero Topic Assignment and Disambiguation}
\label{zta-sec}

In this section, I  briefly describe the {\sc zta} rule proposed
by Walker et al.~and show that discourse coherence indeed tends to be
maintained with the same discourse topic across utterances. In Walker
et al., the centering algorithm specifies  two structures for
centers, namely Cb ({\sc backward-looking center}) and Cf ({\sc
forward-looking centers}), and a set of rules and constraints (See
Walker, Joshi and Prince, this volume).  {\sc forward-looking centers}
are a set of semantic discourse entities associated with each
utterance. The Cf Ranking for Japanese according to discourse salience
is given in (\ex{1}).

\eenumsentence{\item[\ ]
{\sc topic} $>$ {\sc empathy} $>$ {\sc subject}
$>$ {\sc object2} $>$ {\sc object} $>$ {\sc others}
}

The highest ranked member of the Cf list is called the Cp ({\sc preferred
center}). The Cp represents a prediction about the Cb of the following
utterance. The {\sc backward-looking center} is the discourse entity that
the utterance most centrally concerns.  Discourse coherence is
computed with this distinction between looking back to the previous
discourse with the Cb and projecting preferences for interpretation in
subsequence discourse with the Cp.  In other words, the combination of
the Cb and the Cp reflects the coherence of the discourse. The shift
of centers is realized when a new entity is introduced as the Cp.

These interactions of the Cb and Cp are stated as a set of constraints
and rules (Walker, Joshi and Prince, this volume).  What the
constraints and rules amount to is the idea that discourse segments
that continue centering the same entity are more coherent and easier
to process than those that repeatededly shift from one center to
another.  The theory measures coherence by the hearer's inference load
when interpreting a discourse sequence \cite{GJW86,GJW95}.

{\sc zero topic assignment} is a discourse rule which allows a {\it
zero\/} to be interpreted as a {\sc zero topic}.  {\sc zta} is applied
when there is no {\sc continue} transition of the previous center.

\eenumsentence{\item[\ ]
{\bf Zero Topic Assignment} \\ When a zero in $\rm{U_{i+1}}$
represents an entity that was the $\rm{Cb(U_{i})}$, and when no other
{\sc continue} transition is available, that zero may be interpreted
as the {\sc zero topic} of $\rm{U_{i+1}}$.  }

The rule allows a {\it zero\/} that has been the Cb in $\rm{U_{i-1}}$
to continue as the Cp in $\rm{U_{i}}$, even if it appears in a less
salient syntactic position.  It explains why the discourse entity
Hanako, which is realized as the {\sc object2} in (\ex{1})c is
interpreted as the {\sc subject} in (\ex{1})d. Consider again
example (1) repeated here as
(\ex{1}), with the centering data structures:

\eenumsentence{\item[a.]
\shortex{10}{Hanako &wa &siken &o &oete,&kyoositu&ni&modorimasita.}
          {Hanako&{\sc top/subj} &exam &{\sc obj} &finish &classroom &to
&returned}
          {{\it Hanako returned to the classroom, having finished her exam.}}

\begin{tabular}{|lll|}
\hline
{\bf Cb:} & {\sc hanako} & \\
{\bf Cf:} & [{\sc hanako}, {\sc exam}]   & \\ \hline
\end{tabular}

\item[b.]
\shortex{6}{0  &hon &o &locker &ni &simaimasita.}
           {{\sc subj} &book &{\sc obj} &locker &in &took-away}
           {{\it She put her books in the locker.}}

\begin{tabular}{|lll|}
\hline
{\bf Cb:} & {\sc hanako} & \\
{\bf Cf:} & [{\sc hanako}, {\sc book}, {\sc locker}] & {\sc continue} \\ \hline
\end{tabular}

\item[c.]
\shortex{10}{Itumo &no &yooni &Mitiko &ga &0&deki&o&tazunemasita.}
{always& &like &Mitiko&{\sc subj} &{\sc obj2} &result &{\sc obj}
&asked}
{{\it Mitiko, as usual,  asked (Hanako) how she did.}}

\begin{tabular}{|lll|}
\hline
{\bf Cb:} & {\sc hanako} & \\
{\bf Cf1:} & [{\sc hanako}, {\sc mitiko}, {\sc result}] & {\sc zta
continue}  \\
           & {\sc top}, {\sc subj}, {\sc obj} & \\  \hline
{\bf Cf2:} & [{\sc mitiko}, {\sc hanako}, {\sc result}] & {\sc retain}  \\
           & {\sc subj}, {\sc obj2}, {\sc obj}   & \\  \hline
\end{tabular}

\item[d.]
\shortex{10}
{      0 &   0 &zibun&no&tokenakkatta&mondai&o &misemasita.}
{{\sc subj} & {\sc obj2} &self&{\sc gen}&solve-could-not&problem&{\sc
obj}&showed}
{{\it (Hanako) showed (Mitiko) the problems which she could not
solve.}}  \\
{{\it (Mitiko) showed (Hanako) the problems which she could not solve.}}

\begin{tabular}{|lll|}
\hline
{\bf Cb1:} & {\sc hanako} & \\
{\bf Cf1:} & [{\sc hanako},  {\sc mitiko}, {\sc problem}] & {\sc continue}
from Cf1(c)  \\
           & {\sc subj}, {\sc obj2}, {\sc obj} & \\  \hline \hline
{\bf Cb2:} & {\sc mitiko} & \\
{\bf Cf2:} & [{\sc mitiko}, {\sc hanako}, {\sc problem}] & {\sc smooth-shift}
from Cf2(c) \\
           & {\sc subj}, {\sc obj2}, {\sc obj}   & \\  \hline
\end{tabular}
\label{zta-ex-ga}
}

The discourse situation in (\ex{0}) is a case where the hearer may
maintain multiple hypotheses about where the speaker's attention is
directed. There are two assumptions available, the assumption that
{\sc zta} applies and the {\it zero\/} is interpreted as the topic,
versus the assumption that subjects are more highly ranked than
objects on the Cf.  Cf2 of (\ex{0})c is the only Cf possible without
{\sc zta}, and represents a {\sc retain} rather than a {\sc continue}.
By the formulation of the {\sc zta} rule above, {\sc zta} is triggered
here since no {\sc continue} transition is otherwise available. Cf1
represents a {\sc continue} reading due to the {\sc zta} option; {\sc
hanako} can be the Cp even when {\sc mitiko} is realized as the
subject.  This could lead to a potential ambiguity in (\ex{0})d,
because it is possible for a hearer to simultaneously entertain both
of the Cfs in (\ex{0})c. However, the {\sc continue}
interpretation which results from the {\sc zta continue} transition
state is strongly preferred. Walker et al (1994) reported that 28 out
of 34 speakers preferred the {\sc continue} interpretation in
(\ex{0}d); ($ Z = 4.95, p < .001$). The less preferred
{\sc smooth-shift} interpretation would come from the algorithm's
application to Cf2 of (\ex{0})c.

Walker et al.~make a distinction between the notions of {\sc
grammatical topic} and {\sc zero topic}. The grammatical topic is the
{\it wa\/}-marked entity, which is by default predicted to be the most
salient entity. The interaction between the grammatical topic and the
zero topic is observed in (\ex{1}).  Discourse segment (\ex{1}) uses
the {\it wa\/}-marked {\sc np}  instead of the {\sc
ga\/}-marked {\sc np} in the {\sc zta} environment of (\ex{1})c.
Compare the interpretation of  (\ex{1})d with
 (\ex{0})d.

\eenumsentence{
\item[a.]
\shortex{8}{Hanako &wa &siken &o &oete,&kyoositu&ni&modorimasita.}
          {Hanako&{\sc top/subj} &exam &{\sc obj} &finish &classroom &to
&returned}
          {{\it Hanako returned to the classroom, having finished her exam.}}

\begin{tabular}{|lll|}
\hline
{\bf Cb:} & {\sc hanako} & \\
{\bf Cf:} & [{\sc hanako}, {\sc exam}]   & \\ \hline
\end{tabular}

\item[b.]
\shortex{6}{0  &hon &o &locker &ni &simaimasita.}
           {{\sc subj} &book &{\sc obj} &locker &in &took-away}
           {{\it (Hanako) put (her) books in the locker.}}

\begin{tabular}{|lll|}
\hline
{\bf Cb:} & {\sc hanako} & \\
{\bf Cf:} & [{\sc hanako}, {\sc book}]  {\sc continue} & \\ \hline
\end{tabular}

\item[c.]
\shortex{10}{Itumo &no &yooni &Mitiko &{\bf wa} &0&deki&o&tazunemasita.}
{always& &like &Mitiko&{\sc top/subj} &{\sc obj2} &result &{\sc obj}
&asked}
{{\it Mitiko, as usual,  asked (Hanako) how she did.}}

\begin{tabular}{|lll|}
\hline
{\bf Cb:} & {\sc hanako} & \\
{\bf Cf1:} & [{\sc hanako}, {\sc mitiko}, {\sc result}] & {\sc zta
continue}  \\
           & {\sc zero-top}, {\sc top/subj}, {\sc obj} & \\  \hline
{\bf Cf2:} & [{\sc mitiko}, {\sc hanako}, {\sc result}] & {\sc retain}  \\
           & {\sc top/subj}, {\sc obj2}, {\sc obj}   & \\  \hline
\end{tabular}

\item[d.]
\shortex{10}
{      0 &   0 &zibun&no&tokenakkatta&mondai&o &misemasita.}
{ {\sc subj} & {\sc obj2} &self&{\sc gen}&solve-could-not&problem&{\sc
obj}&showed}
{{\it (Hanako) showed (Mitiko) the problems which she could not
solve.}}  \\
{{\it (Mitiko) showed (Hanako) the problems which she could not solve.}}

\begin{tabular}{|lll|}
\hline
{\bf Cb1:} & {\sc hanako} & \\
{\bf Cf1:} & [{\sc hanako},  {\sc mitiko}, {\sc problem}]  {\sc continue}
from Cf1(c)&  \\
           & {\sc subj}, {\sc obj2}, {\sc obj} & \\  \hline \hline
{\bf Cb2:} & {\sc mitiko} & \\
{\bf Cf2:} & [{\sc mitiko}, {\sc hanako}, {\sc problem}]  {\sc smooth-shift}
from Cf2(c)& \\
           & {\sc subj}, {\sc obj2}, {\sc obj}   & \\  \hline
\end{tabular}
\label{zta-ex-wa}
}

The {\it wa} marking has the predicted effect.  Using the grammatical
topic marker {\it wa} in (\ex{0})c dampens {\sc zta} and thus affects
the interpretation of (\ex{0})d, which is now completely
ambiguous. The results of experiments reported in \cite{WIC94} show
that 10 subjects who prefer an interpretation that depends on {\sc
zta} in (\ex{-1}) can no longer get the interpretation in (\ex{0}).
In (\ex{0})d, only 18 out of 34 subjects prefer the {\sc zta continue}
interpretation.  Because the discourse entity realized as the
grammatical topic and indicated by the {\it wa}-marked {\sc np} is the
Cp by default, it is harder to interpret the {\it zero\/} as the
topic.  The situation can be characterized as a case of competing
defaults; some hearers apply the default that the {\it wa}-marked
entity is usually the Cp, and others apply the default that {\sc
continue} interpretations are preferred and that {\it zeros} realize
discourse entities that are ranked highly on the Cf.

When an ambiguity arises from the use of the {\sc wa}-marked {\sc np}
in the {\sc zta} environments as illustrated in the above example, it
is often resolved with additional information provided in the
subsequent discourse. Consider (\ex{1}).\footnote{There is no decisive
proposal how complex sentences should be divided and arranged. In this
study, I simply divide a complex sentence into simplex sentences and
arranged them in serial order. The complex sentences which appeared in
the data consist of coordinations and compounds with temporal adjunct
clauses. A temporal subordinate clause is followed by the main clause
in Japanese, so simple serial ordering normally preserves their
chronological order.}

\eenumsentence{\item[a.]
\shortex{8}{S International&wa&sirikon-varee&ni&kenkyuusyo&o&kaisetusuru.}
          {S International&{\sc top/subj} &silicon
valley&in&laboratory&{\sc obj}&establish}
          {{\it (S International) establishes a laboratory in Silicon Valley.}}

\item[b.]
\shortex{10}{0 &sutaffu&tosite&doobunya&no&keni&hutari&o&sukautosita.}
          {{\sc subj} &staff&as&this-field&{\sc
gen}&authority&2-people&{\sc obj}&recruited}
          {{\it (S International) has recruited two authorities in
the field as a staff.}}

\begin{tabular}{|lll|}
\hline
{\bf Cb:} & {\sc s international} & \\
{\bf Cf:} & [{\sc s international}, {\sc two authorities}]   & \\ \hline
\end{tabular}

\item[c.]
\shortex{10}{Kono&kenkyuusyo&wa&0&saniibeeru&ni&kaisetusi,}
{this&laboratory&{\sc top/obj} &{\sc subj}&Sunnyvale&in&open}
{{\it (S International) will open this laboratory in Sunnyvale,}}

\begin{tabular}{|lll|}
\hline
{\bf Cb:} & {\sc s international} & \\
{\bf Cf1:} & [{\sc s international}, {\sc laboratory}]  {\sc zta continue} &
\\
           & {\sc zero-top}, {\sc top/obj}   & \\  \hline
{\bf Cf2:} & [{\sc laboratory}, {\sc s international}]  {\sc retain} & \\
           & {\sc top/obj}, {\sc subj} & \\  \hline
\end{tabular}

\item[d.]
\shortex{10}{0&0&``Oputo-huirumu-kenkyuusyo''&to&nazukeru.}
{{\sc subj} &{\sc obj}&Opt-film-laboratory&as&name}
{{\it (S International) names (the laboratory) Opt-film Laboratory.}}

\begin{tabular}{|lll|}
\hline
{\bf Cb:} & {\sc s international} & \\
{\bf Cf:} & [{\sc s international}, {\sc laboratory}]  & \\ \hline
\end{tabular}
}

Recall that the {\sc zta} effects are dampened when the grammatical
topic marker {\it wa\/} is used. The third sentence yields the
situation where the zero topic must compete with the grammatical
topic, and the preference for one over the other is hard to determine.
The ambiguity is resolved after processing the fourth sentence,
however, when semantic information about the naming relation is
provided.  In other words, the inference that a newly created thing is
normally given a name, allows the hearer to hypothesize that {\it the
laboratory\/} naturally fills the {\it named\/} slot of the {\it
naming\/} relation.

In sum, these observations support the predictions made by centering
that the preferred interpretation of utterances that contain {\it
zeros} is one in which discourse coherence is maintained. Furthermore,
{\sc zta} allows the hearer to interpret the current utterance as being
highly coherent with the previous utterance. I have also suggested
that in cases where an ambiguity arises because of the use of {\sc zta}, the
speaker will provide additional cues to guide the hearer's
interpretive process.

\section{The Shift of Attentional Focus}
\label{shift-sec}

Now let us consider the prediction that discourse coherence is
maintained even when zeros are used to shift the center. This is the context
in which the Cb in utterance $\rm{U_{i}}$ is not realized as the Cp
(i.e.~the most salient entity in $\rm{U_{i}}$).  A new entity is
introduced as the Cp, and the shift of the speaker's attentional focus
onto this new entity is indicated.  Below, I examine the
interpretation of {\it zero\/}s in  {\sc retain} (discourses (\ex{1}) and
(\ex{2})) and {\sc
rough-shift} (Discourses (\ex{3}) and (\ex{4}))  transitions. After discussing
these examples, I propose some hypotheses about how zeros are interpreted in
these
environments.

In (\ex{1}c) a new center, {\it T co.} is introduced into the discourse
and realized as a  topic, while the old center, {\it the student} is realized
as an object. Thus the center realized by {\it the student} is ranked lower
on the Cf than the center realized by {\it T co.}, but {\it the student} is
still the Cb, so the centering transition
is a {\sc retain}.

\eenumsentence{\item[a.]
\shortex{8}{Gakusei &wa &hurii-daiaru-kaado&de&G-sya&e&denwasureba,}
          {students&{\sc top/subj} &free-dial card&with&G. Company&to&phone}
          {{\it When students call G. Company with the phone card,}}

\item[b.]
\shortex{6}{0  &syuusyoku-zyoohoo&o&muryoo-de&erareru.}
           {{\sc subj} &employment-information&{\sc obj}&free&get-can}
           {{\it (The student) can get employment information free.}}

\item[c.]
\shortex{10}{T-sya&wa&rezyaa-zyoohoo&o&0&fakusimiri&de&teikyuusiteori,}
{T Co.&{\sc top/subj}&leisure info.&{\sc obj}&{\sc obj2}&fax&by&provide}
{{\it T Co. provides leisure information (to student) by fax, }}
}

In (\ex{1}c) a new center, {\it the price} is introduced into the discourse
and realized as a  topic, and the center for {\it the bank} is realized
as a subject. Thus the center for {\it the bank} is ranked lower
on the Cf than the center for {\it the price.}, but {\it the bank} is
still the Cb, so the centering transition
is a {\sc retain}.

\eenumsentence{\item[a.]
\shortex{12}{Saga Ginkoo&wa &gasorin-sutando&de&''banku POS''
saabisu&o&hazimeru.}
          {Saga Bank&{\sc top/subj} &gas station&at&''Bank POS''
service&{\sc obj}& will start}
          {{\it Saga Bank will start ``Bank POS'' service at gas stations.}}

\item[b.]
\shortex{10}{0&kaimono-kayku&ni&kyassyu-kaado&wo&tukatte-morai,}
          {{\sc subj} &shoppers&{\sc obj2}&cash card&{\sc obj}&use-ask}
          {{\it (the bank) asks shoppers to use a credit card,}}

\item[c.]
\shortex{11}{daikin&wa&0&sokuza-ni&kokyaku&no&kooza&kara&hikiotosu.}
          {price&{\sc top/obj}&{\sc subj}&immediately&customer&{\sc
poss}&account&from&draw}
          {{\it (the bank) takes the charge immediately from a customer's
account.}}
}

In (\ex{1}c), the only center that provides a link to the prior discourse is
the center for  {\it the customer}, so that center is the Cb. However  {\it the
customer} is is ranked lower
on the Cf than the center for {\it T. Insurance Co.}, yielding a {\sc rough
shift}
centering transition.

\eenumsentence{\item[a.]
\shortex{8}{S. ginkoo&wa&kinyuu-hosyoo-seido&no&toriatukai&mo&hazimeru.}
          {S. Bank.&{\sc top/subj} &money-insurance-system&{\sc
gen}&handling&{\sc obj}& begin}
          {{\it S. Bank will start to handle a money insurance system
as well.}}

\item[b.]
\shortex{8}{Kokyaku&ga&ittei&ryookin&o&haraeba,}
           {customer&{\sc subj} &certain&fee&{\sc obj}&pay}
           {{\it A customer pays a certain amount of fee,}}

\item[c.]
\shortex{10}{T. Insurance Co.&ga&sono&kinyuu-torihiki&o&0&hosyoosuru.}
{T. Insurance Co.&{\sc subj}&that&money-transaction&{\sc obj}&{\sc
obj2}&insure}
{{\it T Insurance Co.~insures the money transaction {\it (to the
customer\/}).}}
}

In (\ex{1}a), the phrase {\it T. Electron} introduces a center that is
established
as the Cb in (\ex{1}b. Other discourse entities become the Cb in utterances
(\ex{1}d)) to (\ex{1})f, but in (\ex{1}g) the center corresponding to {\it T.
Electron} is realized by a {\it zero}. None of the centers in  (\ex{1})f serve
as an antecedent for this zero, so this is a {\sc rough shift} transition.

\eenumsentence{\item[a.]
\longex{8}{8}{T. Electron&wa&Yamanasi-ken Nirasaki-si&ni&daikibona}
{koozyoo&o&kaisetusuru.}
{T. Electron&{\sc top/subj}&Yamanasi, Nirasaku-city&in&big}
{facotry&{\sc obj}&will built.}
{{\it T.E lectron will open the big factory in Nirasaki City, Yamanasi}}

\item[b.]
(a few sentences about T. Electron)

\item[c.]
\longex{8}{8}{Sinkoozyoo&de&seisansuru&sooti&wa&{\sc te}5000&o}
{seinoo-appu-sita&{\sc rie}-ettingu-sooti.}
{new factory&in&produce-is&devices&{\sc top/subj}&{\sc te}5000&{\sc
obj}}
{power-up-did&{\sc rie}-etching-devices}
{{\it The devices that produced in the new factory are {\sc rie} etching
devices,}} \\
{{\it  more powerful than {\sc te}5000.}}

\item[d.]
\shortex{8}{0&16{\sc mdram}&no&sesan&ni&taioodekiru.}
          {{\sc subj}&16{\sc mdram}&{\sc
gen}&production&{\sc obj2}&cope-with}
          {{\it ({\sc rie} devices) can cope with the production of
16 {\sc mdram}.}}

\item[e.]
\shortex{8}{{\sc dram}&no&syuusekido&ga&takamaruniture,}
          {{\sc dram}&{\sc gen}&integrality&{\sc subj} &increase}
          {{\it As the integrality of {\sc dram} increases,}}

\item[f.]
\shortex{8}{ettyaa&no&zyuyoo&ga&hueru&tame,}
          {etching-devices&{\sc gen}&demand&{\sc subj}&increase&since}
          {{\it The demand of etching devices increases, and hence,}}

\item[g.]
\shortex{8}{0&sinkoozyoo&no&seisan&ni&humikitta.}
          {{\sc subj}&new facility&{\sc gen}&production&{\sc obj2}&decided}
          {{\it (T. Electron) decided to begin the production in the
new facility.}}
}

Note that the interpretation of {\it zeros} is not particularly problematic in
the case
of {\sc retain}; although the Cb is shifting the antecedent for the
{\it zero} is a center from the previous utterance. Furthermore, in some
cases,  the {\sc retain} transitions may have a {\sc
zta continue} option. However in  the
{\sc rough-shift} transition, no local antecedent of a {\it
zero\/} is available and a center shift is forced. In this second case,
the {\it zero\/}'s antecedent is not in the immediately
preceding utterance, but must be realized in prior utterances. These
cases have been  called {\sc return pops} or {\sc focus
pops} in the literature \cite{Reichman85,PS84,GS86}. See also (Walker,
this volume).

\begin{figure}[htb]
\begin{tabular}{|r|c|c|c|}
\hline
& {\sc lexical}  & {\sc tense \&} & \\
& {\sc semantics} & {\sc aspect} & {\sc agreement}\\ \hline
 & & & \\
{\sc rough-shift} with {\it zeros} & 20 & 6 & 2 \\ \hline
\end{tabular}
\caption{{\bf Disambiguation Features for Rough-Shift}}
\label{lex-fig}
\end{figure}

If discourse coherence is to be maintained, it seems clear that there
must be other cues that are used to preserve coherence and resolve
{\it zero\/}s appropriately. This prediction has turned out to be
correct. To test the hypothesis that shifting centers are associated
with contextual factors that facilitate transitions, such as lexical
semantics, agreement information and tense and aspect, all the rough
shifts in the corpus (23 of them) wre coded for these features. The
results are given in figure \ref{lex-fig}.\footnote{The total number
of the table exceeds the total number of 23 occurrences of the {\sc
rough-shift} transition with zeros. This is due to the fact that there
are some cases where two features (i.e.~lexical semantics and tense)
are employed at the same time.}  Below I illustrate the role of these
factors in interpreting {\it zeros} when the center shifts with
representative examples from the corpus.

\subsection{Interaction with lexical semantics}

Let us take a look at the  discourse in (\ex{0}).  The
appropriate interpretation of the {\it zero\/} in the last sentence is
constrained by the semantic restriction assigned to the arguments of
verb `decide'.  No entity in (\ex{0})f can be a potential
antecedent, and the {\it zero\/} must be resolved to a discourse entity
expressed in the previous utterances of the text. In this case, it
goes back to the utterance where {\it T. Electron\/} is
available.\footnote{The part indicated by italics is the segment given
in (\ex{0}).}

\eenumsentence{\item[\ ]
{\it {\bf T. Electron} will open the biggest factory in Nirasaki City,
Yamanasi.} (T. Electron) will build (the factory) in the company
property adjacent to its General Laboratory. (T. Electron) will
provide a big-scale clean room, and produce etching devices which can
deal with 16M bit dynamic {\sc ram}. The total investment amounts to 5
billion yen and the construction starts this fall. It is expected that
(the factory) will start operation in a year later. {\it The
devices produced in the new factory are {\sc rie} devices, more
powerful than {\sc te5000}.  ({\sc rie} devices) can cope with the
production of 16{\sc mdram}.  As the integrality of {\sc dram}
increases, the demand of etching devices increases, and hence, (T.
Electron) decided to begin the production in the new facility. } }

If we assume that the antecedent of a {\it zero\/} is any of the
centers introduced in the previous discourse, the interpretation of
the last sentence would be ambiguous; there are multiple potential
candidates even if lexical information is brought to bear.  Note that
{\it Nirasaki City\/} and {\it General Laboratory\/} are semantically
legitimate antecedents of the missing subject of the {\it
deciding\/}-situation described by the last sentence. The
uncontroversial interpretation with {\it T. Electron\/} as the
antecedent suggests that a discourse entity that has not been
previously realized as the Cb {\bf cannot} be interpreted as the
cospecifier of a {\it zero\/}.

Discourse coherence can be maintained by an inference process based on
the lexical semantics, but the preferred interpretation is not always
computed by a inference process purely driven by the underlying
semantics. Instead, discourse information such as attentional focus
and salience provides constraints on the application of information
from lexical semantics.  This interaction is key for enhancing
centering by incorporating disambiguation information from other
sources.

This claim is further supported by the observation in (\ex{1}). If we
assume that the antecedent of a {\it zero\/} can be any of the
entities that were previously realized in a discourse, nothing stops
the {\it zero\/} in the third utterance from taking {\it doosya\/}
(`the company') in the first utterance as its antecedent since this
would yield a semantically plausible {\sc rough-shift}
interpretation. However, this interpretation is never preferred over
the interpretation obtained by a more highly ranked centering
transition. That is, no interpretation based on lexical semantics is
preferred to an interpretation that is ranked higher in terms of
centering transitions. The preferred interpretation according to the
centering rules cannot be overridden unless this interpretation is
semantically anomalous.

\eenumsentence{\item[a.]
\shortex{8}{doosya&wa  &15-dai &no &hanbai& o& mikondeiru.}
          {company&{\sc top/subj} &15-piece &{\sc gen}& sales &{\sc obj}
&anticipate}
          {{\it The company anticipates the sales of 15 machines.}}

\begin{tabular}{|lll|}
\hline
{\bf Cb:} & {\sc company} & \\
{\bf Cf:} & [{\sc company}, {\sc sales}]   & \\ \hline
\end{tabular}

\item[b.]
\shortex{6}{{\sc cvd}-sooti  &wa &{\sc ceraus}.}
           {{\sc cvd}-device&{\sc top/subj} &{\sc ceraus}}
           {{\it The {\sc cvd} device is (called) {\sc ceraus}.}}

\begin{tabular}{|lll|}
\hline
{\bf Cb:} & {\sc company} & \\
{\bf Cf:} & [{\sc cvd-device}, {\sc ceraus}] & {\sc retain} \\ \hline
\end{tabular}

\item[c.]
\shortex{10}{0 &maruti-tyenbaa-hoosiki& o &saiyoo.}
{{\sc subj} &multi-chamber system& {\sc obj} & adopt}
{{\it (CVD-device) adopts a multi-chamber system.}}

\begin{tabular}{|lll|}
\hline
{\bf Cb1:} & {\sc cvd-device} & \\
{\bf Cf1:} & [{\sc cvd-device}, {\sc system}]  {\sc smooth-shift} & \\
           &  {\sc subj}, {\sc obj} & \\  \hline
{\bf Cb2:} & {\sc cvd-device} & \\
{\bf Cf2:} & [{\sc company}, {\sc system}]  {\sc rough-shift} & \\
           &  {\sc subj}, {\sc obj} & \\  \hline
\end{tabular}

\item[d.]
\shortex{5}
{      0 & tahaisen-maku &ni &taioo-dekiru.}
{ {\sc subj} & multi-wired film &{\sc obj2} & deal-can}
{{\it (CVD-device) can deal with multi-wired films.}}

\begin{tabular}{|lll|}
\hline
{\bf Cb1:} & {\sc cvd-device} & \\
{\bf Cf1:} & [{\sc cvd-device}, {\sc films}]  {\sc continue} & \\
           &  {\sc subj}, {\sc obj2} & \\  \hline
{\bf Cb2:} & {\sc company} & \\
{\bf Cf2:} & [{\sc company},  {\sc films}]  {\sc smooth-shift} & \\
           &  {\sc subj}, {\sc obj2} & \\  \hline
\end{tabular}
}

The lexical semantics of the verb {\it saiyoo\/} (`adopt') in
(\ex{0})c would not block {\it the company\/} in (\ex{0}a) being
realized as their subject.  For instance, both `{\it The {\sc cvd}-device
adopts a multi-chamber system}' and `{\it the company adopts a
multi-chamber system\/}' are reasonable readings of (\ex{0})c.  However,
the Cf2 reading, which is obtained on the basis of lexical
semantics and yields the {\sc rough-shift} transition, is not
preferred to the Cf1 {\sc smooth-shift} reading.  The preference
assigned to (\ex{0}c) based on centering transitions is seen in
(\ex{0})d.

The verb {\it taioo} in (\ex{0}d) means `answer' or `response' when
the human being or the organization is the subject, and it normally
takes an abstract noun such as as `demand', `a political crisis' as
its object. The verb also takes the non-agentive entity as the subject
and means its applicability to some other object expressed in the
non-subject position. The missing subject of the sentence in
(\ex{0})d, which has a concrete object in the object2 position,
therefore naturally refers to {\it the {\sc cvd} device\/} rather than
{\it the company\/}, meaning that the {\sc cvd} device is applicable
to handle  multi-wired films. The preferred interpretation of
(\ex{0})d thus supports the preference computed in utterance (\ex{0})c
based on the centering transitions; the interpretation, which preserves
discourse coherence between discourse segments, is the one most
preferred.

Thus, lexical semantics can be used to resolve the
interpretation of {\it zero\/}s, as long as its interaction with discourse
information
about attentional state is taken into consideration.

\subsection{Interaction with tense and aspect}

It is not always the case that lexical semantics provides a cue.
Observe the following examples.

\eenumsentence{\item[a.]
\shortex{8}{T. Electron&wa  &hiitaa-koozyoo&no&kensetu&ni&tyakusyusita.}
          {T Electron&{\sc top/subj} &heater factory &{\sc gen}&
construction &{\sc obj} &began({\sc past})}
          {{\it T. Electron began the construction of its heater factory.}}

\item[b.]
\shortex{10}{koremade&0&kyoodaigaisya&kara&kyookyuu&o&uketeita&ga,}
           {by now&{\sc subj}&brother-company&from&supply&{\sc
obj}&recieved&but}
           {{\it By now (T. Electron) has been receiving the supply
from its brother company,}}

\item[c.]
\shortex{10}{0 &zisya-seisan&ni&kirikaeteiku.}
{{\sc subj} &self-production&{\sc obj2} & introduce}
{{\it (T. Electron) will introduce self-production.}}

\item[d.]
\shortex{10}
{Hiitaa-koozyoo&wa&0&itagane-koozyoo&ni&rinsetusite&kensetusuru.}
{Heater factory& {\sc top/obj} & {\sc subj}&steel
factory&to&adacent&construct.}
{{\it (T.Electron) is constructing the heater factory next to the steel
factory.}}

\item[e.]
\shortex{10}
{0&hiraya-date&de, &yukamenseki&658 heihoo-meetoru.}
{{\sc subj}&one-story&is&floor space&658 square meter}
{{\it (The heater factory) is one-story building with the floor space
of 685 square meter.}}

\item[f.]
\shortex{10}
{0&{\sc cvd}-sooti-yoo hiitaa&o&seisansuru.}
{{\sc subj}&{\sc cvd}-device-for heater&{\sc obj}&produce}
{{\it (The heater factory) will produce heaters for {\sc cvd}-devices.}}

\item[g.]
\shortex{10}
{Toosigaku&wa&2-oku 8-sen man yen&da.}
{investment-money&{\sc top/subj}&280 million yen&is}
{{\it The investment money amounts to 280 million yen.}}

\begin{tabular}{|lll|}
\hline
{\bf Cb:} & {\sc heater factory} & \\
{\bf Cf:} & [{\sc investment}, {\sc 280 million yen}]  {\sc retain} & \\
           &  {\sc subj}, {\sc comp} & \\  \hline
\end{tabular}

\item[h.]
\shortex{12}
{0&san'nin&no&gizyutusya&o&Sagami&ni&gizyutusyuutoku&tame&hakensita.}
{{\sc subj}&three&{\sc gen}&technician&{\sc obj}&Sagami&to&technical
training&for& sent}
{{\it (T. Electron) sent three technicians to Sagami for technical training.}}

\begin{tabular}{|lll|}
\hline
{\bf Cb1:} & {\sc investment money} & \\
{\bf Cf1:} & [{\sc heater factory}, {\sc technician}]  {\sc rough-shift} & \\
           &  {\sc subj}, {\sc obj} & \\  \hline
{\bf Cb2:} & {\sc investment money} & \\
{\bf Cf2:} & [{\sc T. Electron}, {\sc technician}]  {\sc rough-shift} & \\
           &  {\sc subj}, {\sc obj} & \\  \hline
\end{tabular}
}

No entity in (\ex{0})g is suitable as an antecedent of the {\it zero} in
(\ex{0})h -- {\it the investment money\/} is never interpreted as the
{\it sender\/} in (\ex{0}).\footnote{Here  {\it heater
factory\/} and {\it T. Electron\/}, realized in the previous discourse
segments, are potential antecedents of the {\it zero\/} because they
both meet the constraints on the antecedency of zeros and the
semantics of the verb. However, the following alternative analysis
would be possible.  The introduction of a new entity, {\it
tossigaku\/} (`investment money') in (\ex{0})g may indicate that this
entity is associated with an entity that has been already introduced
in the discourse. That is, we can assume that there is functional
dependency relation between {\it heater factory\/} and {\it investment
money\/}; {\it investment money\/} is the money for establishing the
heater factory.  In other words, the heater factory might be
implicitly realized in (\ex{0})h though it is not overtly
expresed. More research should be done to formalize when such an
implicit relation is realized.  A statistical measure of cooccurrence
of {\sc np}s may be useful to identify potential attributes associated
with an entity. For instance, a company may have attributes {\sc
name}, {\sc location}, {\sc owned-by}, {\sc product}, {\sc net-worth},
{\sc nationality} and {\sc the number of employees} and so on.} That
is, the {\sc rough-shift} transition is forced to make sense out of
(\ex{0})h and the {\it zero\/} looks for its potential antecedent in
the previous utterances. There are two entities whose semantics is
compatible with what the verb of the sentence requires as its
argument. That is, it is both plausible to say that `{\it The heater
factory (as an organization, though the construction of its building
has not been completed) sent technicians\/}' as well as `{\it T.
Electron sent technicians\/}'.  However, the second reading is more
preferred.  I assume that the shift is supported by
the use of the past tense:\footnote{Tense in Japanese is realized as
the morpheme attached to the verb stem. In general, for the
[$-$stative] verbs, the simple present (or non-past) tense is marked
with {\it -u\/}, while the simple past (or perfect) tense with {\it
-ta\/}.  The present tense form of [$-$stative] verbs usually refers
to future time unless they represent habitual or generic actions, in
which case they refer to present time (Kuno 1973).  The past form
represents an action that has been completed or executed at
reference time.} the attentional focus in (\ex{0})h returns to an
event which has been completed at the time of the utterance. Note that {\it T
Electron} has been mentioned as an entity which conducted some past
action at the beginning of the text.

The example illustrates how inference based on temporal/aspectual
information can be used to resolve ambiguity when no local constraints are
available.  They are used to control the flow of information,
indicating the shift of the reference point in describing events. In
other words, temporal/aspectual coherence participates in an inference
system to maintain non-local coherence and it provides a cue to
identify discourse structure segments and their non-local hierarchical
relations in discourse.

\subsection{Interaction with agreement}

The third strategy to maintain discourse coherence is one that
uses different types of  agreement information in order to elicit adequate
inference and eliminate an undesired potential interpretation. Consider
example (\ex{1}).

\eenumsentence{\item[a.]
\shortex{8}{S. Metal&wa  &zisedaigata&ettingu-sooti&o&kaihatu,}
          {S. Metal&{\sc top/subj} &next-generation-type&etching-device&{\sc
obj} &develop}
          {{\it S. Metal has developed next-generation type etching devices,}}

\item[b.]
\shortex{10}{0&kotosi&kara&honkakuteki-na&maaketingu&o&hazimeteiru.}
            {{\sc subj}&this year&from&full-scale&marketing&{\sc obj}&begin}
           {{\it (S. Metal) has started full-scale marketing this year.}}

\item[c.]
(a few sentences about {\it the etching device\/})

\item[d.]
\shortex{10}{{\sc cvd}-sooti&wa&kore&ni&tuzuku&mono&de,}
{{\sc cvd}-device&{\sc top/sub} &this&{\sc obj2}&follow&thing&be}
{{\it {\sc cvd} devices are the thing that will follow this (i.e.~etching
devices).}}

\item[e.]
\shortex{10}
{habahiroi&zyuyoo&ga&kitaisareteiru.}
{wide&demand&{\sc subj} &is-expected}
{{\it Wide range of demand is expected.}}

\item[f.]
\shortex{10}
{0&{\bf tomoni}&{\sc ecr}&o&riyoositeori,}
{{\sc subj} &both&{\sc ecr}&{\sc obj}&use}
{{\it (CVD devices and etching devices) both use ECR,}}
}

The adverb {\it tomoni\/} (`both') in (\ex{0})f indicates that the
unexpressed subject in the utterance refers to a set of two entities.
Considering the previous discourse, we see that the entities which are
of the same type and can form a set in this discourse segments are
{\it etching devices\/} and {\it CVD devices\/}.  Without this
quantifier-like adverb, the {\it zero\/} could refer to {\it
S. Metal\/}, which is an legitimate antecedent of the {\it zero\/} by
itself.

Although the language does not mark number distinction (i.e.~singular
vs. plural) on nouns, classifiers are used when the number or the
quantity does matter; {\it two cups of tea\/}, {\it 3 individuals of
professors\/}, {\it 5 things of apples\/} and so on.  Expressions which
are sensitive to  number as in (\ex{0}) thus can be used
to make an adequate grouping among the entities in a discourse
and prune an undesired interpretation which is otherwise predicted or
never eliminated by  basic discourse coherence principles.

\subsection{Summary}

In conclusion, a shift of centers occurs only when such an intended
interpretation is well supported by other contextual information, so
that the speaker's intention is rarely misinterpreted.  If the speaker
is concerned that her utterance might be misinterpreted as a
consequence of shifting the topic, she always has an alternative to
express the intended new topic overtly as I originally hypothesized in
(3) above.  However, constraints that arise from lexical semantics and
the event structure appear to be readily available cues that the
hearer can use to interpret zeros with nonlocal antecedents. In the
following section, I will discuss how these observations can be
incorporated into centering, and go some way towards integrating
centering with a model of global discourse structure (cf.
\cite{Hobbs85a,PS84,Reichman85,GS86}.

\section{Integrating Centering  and Global Coherence}
\label{integ-sec}

Although our initial hypothesis was that zeros would not be used to
shift centers, we saw above that this often happens in naturally
occurring discourse.  The relevant numbers are repeated with the
definition of the various centering transitions in figure
\ref{trans-state2-fig}.
\footnote{Note that Figure \ref{trans-state2-fig} table shows the
frequency of each transition when an utterance contains at least one
{\it zero\/}.}

\begin{figure}[htb]
\begin{tabular}{|r|c|c|}
\hline & &\\
& $\rm{Cb(U_i) = Cb(U_{i-1})}$ & $\rm{Cb(U_i) \neq Cb(U_{i-1})}$
\\  \hline
 & & \\
$\rm{Cb(U_i) = Cp(U_i)}$ & {\sc continue 76} & {\sc smooth-shift 34} \\ \hline
 & & \\
$\rm{Cb(U_i) \neq Cp(U_i)}$ & {\sc retain 3} & {\sc rough-shift 23} \\ \hline
\end{tabular}
\caption{\bf Distribution of Centering Transitions  with Zeros}
\label{trans-state2-fig}
\end{figure}

In the current algorithms of Centering Theory \cite{BFP87,WIC94},
interpretations are determined by the Cb and Cf in $\rm{U_{i-1}}$ and
$\rm{U_{i}}$ (i.e.  local discourse entities). However, the
observations above suggest that the theory must support an algorithm
for accessing non-local antecedents when a {\sc rough-shift}
transition occurs and a shift to a non-local center is detected (cf.
\cite{Sidner83a}).

In order to capture global coherence, another center data structure
must be added to keep track of the Cbs introduced in the
previous utterances. My data shows that {\it zeros\/} in {\sc
rough-shifts\/} realize discourse entities that were previously
realized as the $\rm{Cb(U_{i-n})}$: there are {\bf no} cases where a
{\it zero\/} realizes a discourse entity that was previously a non-Cb.
Thus I propose that what is needed is a Cb retrieval mechanism of some
type to model the cases where a zero is resolved to a discourse entity
that was an earlier center.

This Cb retrieval mechanism could be based on the stack mechanism of
\cite{Sidner83a,GS86}, or the cache mechanism proposed in
\cite{Walker96b} and discussed in (Walker, this volume). Since I have
no evidence that anything more powerful than a {\sc list} is required,
the proposed algorithm is to search a linearly ordered {\sc list} of
former Cbs, ordered by recency. In all the cases in my data, it is
sufficient to search back through a list of former Cbs ordered by
recency and choose as the antecedent of the zero the first such Cb
that is semantically compatible with the requirements of the
zero. This mechanism for computing global coherence must interact with
the centering algorithm for local coherence in such a way that the
former is activated when the latter fails.  The condition may be
stated as follows.

\eenumsentence{\item[\ ]
If $\rm{Cb(U_i)\neq Cp(U_i)}$, then take $\rm{Cb(U_m)}$ which is an
element of $\rm{M}$ (i.e.~$\rm{Cb(U_m)\in M}$) where $\rm{M}$ is
a list of $\rm{Cb(U_{1\ldots(i-1)})}$ which satisfies the inference process. }

When local coherence is not observed and the shift of the center is
forced, the list of the Cbs of the previous discourse, $\rm{M}$, is
searched, and each proposed Cb is checked against an inference process
based on lexical semantics and tense and aspect information, to
determine its adequacy.  The algorithms to refer to the global
discourse may be sketched as in (\ex{1}).

\eenumsentence{\item[\ ]
When a Cb shift is detected (i.e.~${\rm{Cp(U_i)\neq
Cb(U_i)}}$):
\begin{enumerate}
\item {\bf Local Coherence Check}:\\
   \ \ \ \ \ {\it if\/} {\sc retain} and no {\sc zta-continue}
  is available, go to Global Coherence Check\\
  \ \ \ \ \ {\it if\/} {\sc rough-shift},
            go to Global   Coherence Check \\
  \ \ \ \ \ {\it else\/} return to Centering algorithm \\
\item{\bf Global Coherence Check}:\\
    \ \ \ \ \ Take a $\rm{Cb(U_m)}$ on the $\rm{Cb}$ list, and
  (e.g.~$\rm{Cp(U_m)\in M}$) \\
    \ \ \ \ \ Employ inference systems  \\
\item[3.] {\bf Decision: }\\
      \ \ \ \ \ {\it if\/} the interpretation $\rm{Cp(U_i)=Cb(U_m)}$ is
  acceptable, return to Centering algorithm\\
      \ \ \ \ \ {\it else\/}
            return to Global Coherence Check and try the
            next Cb on the Cb list
\end{enumerate}
}

\section{Discussion}
\label{discuss-sec}

In this chapter, I discuss issues that centering theory needs to
address in order to model discourse coherence in a larger context. I
argue that the use of {\it zeros} to realize previous Cbs in {\sc
retain} and {\sc rough-shift} centering transition states indicates
that coherence information provides constraints on inferential
processes. Future work must integrate these observations with other
studies on shifting centers.  The data examined here show that lexical
semantics as well as temporal/aspectual information are used to create
links between non-local utterances, and that Centering theory can be
extended to compute non-local discourse coherence as long as it
incorporates a richer semantic representation of utterances. I propose
that the combination of the centering algorithm with a global Cb list
captures some aspects of global coherence, without introducing a
completely different module.  This kind of mechanism suggests that it
might be possible to use Centering as a part of an algorithm for
inferring discourse structure.

\end{document}